\documentclass[5p]{elsarticle}

\usepackage{hyperref}

\usepackage{graphicx}
\usepackage{subfigure}
\usepackage{amsmath,amssymb}
\usepackage{multirow}
\usepackage{algorithm}
\usepackage[]{algorithmic}
\usepackage{siunitx}

\usepackage[T1]{fontenc}
\usepackage[utf8]{inputenc}
\usepackage{soulutf8}
\usepackage[super]{nth}

\usepackage{color}
\usepackage{soul}

\usepackage[capitalise, noabbrev]{cleveref}

\journal{Microelectronics Reliability}

\begin{document}

\begin{frontmatter}

        \title{Non-destructive Fault Diagnosis of Electronic Interconnects by Learning Signal Patterns of Reflection Coefficient in the Frequency Domain
}

		\author[UoS]{Tae Yeob Kang}
		\author[AIMMO]{Haebom Lee}
		\author[DFKI,TUKaiserslautern]{Sungho Suh\corref{mycorrespondingauthor}}
        \cortext[mycorrespondingauthor]{Corresponding author}
        \ead{sungho.suh@dfki.de}
        \address[UoS]{Department of Mechanical Engineering, The University of Suwon, Hwaseong, Republic of Korea}
        \address[AIMMO]{AIMMO, Seoul, Republic of Korea}
        \address[DFKI]{German Research Center for Artificial Intelligence (DFKI), Kaiserslautern, Germany}
		\address[TUKaiserslautern]{Department of Computer Science, RPTU Kaiserslautern-Landau, Kaiserslautern, Germany}

		\begin{abstract}
Fault detection and diagnosis of the interconnects are crucial for prognostics and health management (PHM) of electronics. Traditional methods, which rely on electronic signals as prognostic factors, often struggle to accurately identify the root causes of defects without resorting to destructive testing. Furthermore, these methods are vulnerable to noise interference, which can result in false alarms. To address these limitations, in this paper, we propose a novel, non-destructive approach for early fault detection and accurate diagnosis of interconnect defects, with improved noise resilience. Our approach uniquely utilizes the signal patterns of the reflection coefficient across a range of frequencies, enabling both root cause identification and severity assessment. This approach departs from conventional time-series analysis and effectively transforms the signal data into a format suitable for advanced learning algorithms. Additionally, we introduce a novel severity rating ensemble learning (SREL) approach, which enhances diagnostic accuracy and robustness in noisy environments. Experimental results demonstrate that the proposed method is effective for fault detection and diagnosis and has the potential to extend to real-world industrial applications.
		\end{abstract}







\end{frontmatter}


\section{Introduction}
\label{sec:introduction}
Despite the continuous demand for higher computing power, the performance of essential building blocks, including transistors, memories, and processors, is plateauing. Following the aggressive downscaling of advanced integrated circuits (ICs), electronic interconnects have become the bottleneck for the reliability and performance of entire electronic systems \cite{1}. Defects in these interconnects can lead to system failures, decreased performance, and reduced product lifespan, making their early detection and diagnosis critical. Regardless of where or when defects in the interconnects occur, they should be detected early on, and the corresponding parts replaced. However, existing diagnostic methods for electronic interconnects face significant challenges. There has been a growing interest in monitoring the ongoing health of products and systems to predict failures before a catastrophe. Consequently, prognostics and health management (PHM) technologies have been developed \cite{2}. PHM methods can predict failure, diagnose defects, and eventually improve system quality as well as extend system life. Determining the cause of the failure is essential to enabling forecasted maintenance and reducing downtime. Extensive studies have been conducted on the fault diagnosis of mechanical systems by processing vibration signals \cite{tama2023recent,gao2015survey}. 
In contrast, for electronic interconnects, there is a lack of research on non-destructive methods that can identify both the severity and root causes of defects.
Rather, existing diagnosis methods for electronics require destructive testing to determine the root causes of the defects. Furthermore, previous defect detection methods utilizing DC resistance \cite{3,4}, time domain reflectometry (TDR) \cite{5,tdr_new,tdr_new2,tdr_new3,tdr_new4}, Scattering parameter (S-parameter) \cite{6,7,8,9}, radio frequency (RF) impedance \cite{10, 11, 12}, and digital techniques \cite{13,14} are vulnerable to noise, eventually leading to false alarms. 
\begin{figure*}[!t]
    \begin{center}
        \includegraphics[width=\linewidth]{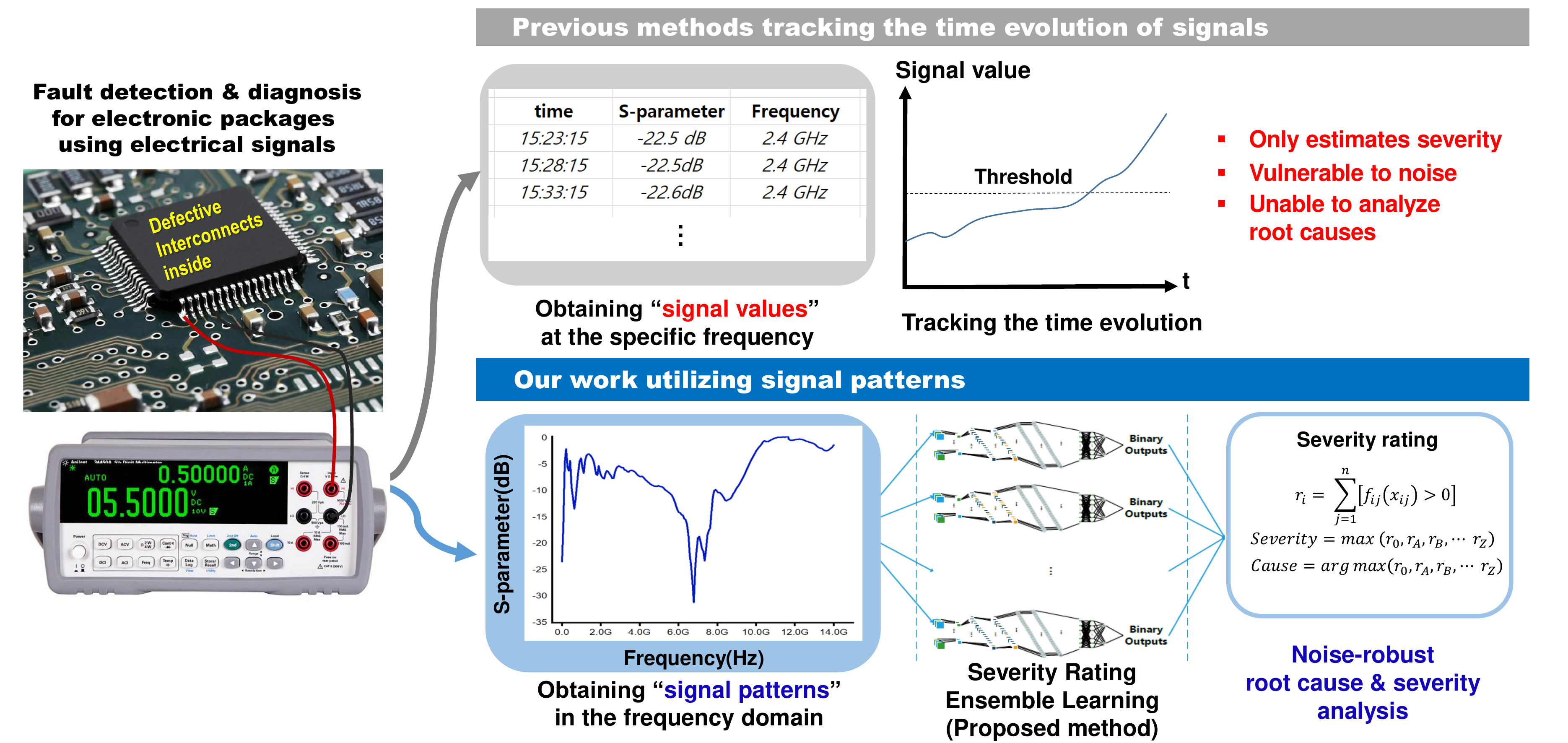}
        \caption{Comparison between our work and previous studies for fault detection and diagnosis of electronic packages using electrical signals}
        \label{fig:comparison}
    \end{center}
    \vspace{-5mm}
\end{figure*}
These methods rely on electronic signals at designated operating frequencies (or in the case of DC resistance, at 0 Hz), which limits their effectiveness. While methods tracking the time evolution of the parameters can estimate the severity of defects effectively, their one-dimensional (1D) trends are insufficient to distinguish the root causes of the defects.
Nevertheless, previous fault detection methods for electronic packages using electronic signals have not yet utilized the signal patterns obtained in a full range of operating frequencies. Instead, often discarded were outliers and irregularities in the signals from defective interconnects. For normal interconnects, certain signal patterns of electronic signals reveal key features, serving as indicators of interconnect characteristics \cite{15}. This observation suggests that the full signal patterns, including those previously considered as outliers, may contain valuable information about defects. Similarly, learning to recognize signal patterns of defective interconnects would enable the extraction of valuable information about the defects.

Therefore, there is a critical need for a diagnostic approach that can non-destructively detect defects, determine their root causes, and operate effectively in noisy industrial environments. The motivation behind our research is to address this need by developing a method that leverages the full frequency range of reflection coefficient signal patterns, combined with robust machine learning techniques, to provide comprehensive fault diagnosis of electronic interconnects.

To address these issues, we propose a novel approach based on discovering specific patterns of electronic signals that provide information about both the severity and cause of defects. We focus on the reflection coefficient signal patterns in the frequency domain, which are distinct from traditional time-series analysis methods for fault diagnosis of electronic packages. The reflection coefficient is an electronic parameter that describes how much of a voltage wave is reflected by an impedance discontinuity in the transmission medium. It is also reported that the coefficient is sensitive to defects in the interconnects\cite{5,tdr_new}. By analyzing the reflection coefficient across a full range of frequencies, we aim to capture comprehensive signal patterns that can reveal detailed characteristics of defects. In this study, we gather the reflection coefficient signal patterns of electronic interconnects with various causes and severity levels of defects. The electronic interconnects serve as representative forms of electronic interconnection  \cite{16}. Then, by applying machine learning (ML) and deep learning (DL) techniques, we diagnose the defects only with the signal patterns of the reflection coefficient. The patterns provide an opportunity to apply the learning algorithms to root cause analysis, while the time series data of electronic signals only estimate the severity of defects.

Furthermore, we introduce a novel severity rating ensemble learning (SREL) approach for fault detection and diagnosis of electronic interconnects, which fully utilizes characteristics of the reflection coefficient signal patterns. Although the conventional ML and DL algorithms with the reflection coefficient patterns provide effective performance in fault diagnosis of electronic interconnects, the conventional algorithms remain susceptible to noise. This vulnerability is a significant concern, particularly given the prevalence of noise in industrial environments. The proposed SREL method improves diagnostic performance and robustness to noise in real industrial environments. Combining the advantages of the reflection coefficient patterns and the SREL method, we aim to significantly advance the field of non-destructive fault diagnosis, improving the reliability and performance of electronic systems in industrial environments. In the experimental results, we demonstrate that the signal patterns of the reflection coefficient extract distinguishable features for different defect states and the proposed SREL method outperforms conventional ML and DL methods, particularly under noisy conditions. As shown in \cref{fig:comparison}, our method achieves early detection, noise-robustness, and non-destructive analysis of root causes, overcoming the limitations of previous studies for fault diagnosis using electronic signals.

The main contributions of this study are as follows:
\begin{itemize}
    \item We address the critical need for non-destructive, noise-robust fault diagnosis methods by obtaining signal patterns of the reflection coefficient according to various defect states of electronic interconnects, demonstrating that these patterns enable both early and accurate diagnosis. 
    \item We show that the signal patterns in the frequency domain are capable of root cause analysis, whereas the previous methods using electronic signals such as DC resistance, TDR, and S-parameters at designated frequencies are not. 
    \item We demonstrate that the signal patterns are effective features for learning algorithms by conducting dimension reduction on the patterns and providing diagnostic results with the conventional ML and DL methods. 
    \item We propose a novel severity rating ensemble learning (SREL) approach that fully exploits the characteristics of the signal patterns so that we enhance the diagnostic performance and noise robustness.     
\end{itemize}

The remainder of this paper is organized as follows. \cref{sec:relatedworks} presents a comprehensive review of related works in the field. \cref{sec:method} details the proposed approach, including the reflection coefficient patterns and the SREL approach. In \cref{sec:ExperimentalDesign}, we describe the experimental design used to evaluate the effectiveness of our approach. \cref{sec:experimentalResults} presents the quantitative and qualitative experimental results. Finally, \cref{sec:conclusion} concludes this paper.

\section{Related Works}
\label{sec:relatedworks}
Previous studies for fault detection and diagnosis of electronic interconnects obtained the signal values at designated operating frequencies and tracked the time evolution. This process can estimate the severity of defects effectively, but it is unable to determine root causes without the disassembly of electronic packages and destructive testing. DC resistance, S-parameter, RF impedance, digital signals, and TDR are the electronic parameters that have been utilized in that manner. The DC resistance measurement method has been widely used for reliability monitoring of electronic interconnects because of its simplicity and convenience. The DC resistance responds to a short or an open-state conductor quite well. However, it is not well-suited to indicate the evolution of defects \cite{kwon2010detection}. To overcome this limitation, studies have suggested various electronic parameters at high frequencies as indicators of defect growth. The electronic signals at the high frequencies capitalize on the skin effect to detect defects at an early stage \cite{5}.

S-parameters have demonstrated superior sensitivity compared to DC resistance as defects progress \cite{6}. Researchers, such as Putaala et al. \cite{7}, monitored S-parameters during temperature cycling tests on Ball Grid Array (BGA) components, revealing qualitative changes in S-parameters correlating with component degradation, unlike the negligible change in DC resistance. Foley et al. \cite{8} proposed void detection in transmission lines by monitoring changes in the leakage conductance parameter derived from S-parameter measurements. RF impedance, sensitive to surface defects due to the skin effect, has shown promise in detecting electronic component defects \cite{9}. Mosavirik et al. \cite{mosavirik2023impedanceverif} proposed an on-chip impedance sensing approach for detecting various classes of tamper events in cryptographic devices.  Digital signal degradation resulting from physical circuit damage has also been explored \cite{13}, leading to on-chip health sensing methods for interconnect degradation detection \cite{14}. TDR, a time domain parameter of the reflection coefficient, has been widely applied for fault detection in wiring networks \cite{tdr_new3,tdr_new4}. While these methods offer valuable insights, they exhibit dependency on operating frequencies, making them susceptible to noise and false alarms. The robustness of fault diagnosis methods to noise in high-frequency parameters remains an understudied aspect. 

Applications of artificial intelligence (AI) primarily involved ML regression techniques \cite{11, 18, 19} for electronics fault detection, focusing on remaining useful life estimation. 
         Recent works extend the application of ML and DL techniques in diverse contexts. Chien \cite{kao2023deep} proposed a DL-based fault diagnosis framework for semiconductor backend processes, demonstrating its effectiveness in predicting maintenance needs. Wächter et al. \cite{wachter2022using} used ML for anomaly detection on a system-on-chip under gamma radiation. Bhatti et al. \cite{bhatti2023neural} presented a neural network-based signal integrity assessment model for on-chip interconnects in integrated circuits. Fang et al. \cite{fang2023self} introduced a prior knowledge-guided teacher-student model for self-supervised intermittent fault detection in analog circuits. Also, the idea of defect identification using frequency or phase measurement and machine learning methods is established in optical metrology \cite{pandey2023non,narayan2023deep,pandey2022subspace,vishnoi2021automated}.        

Despite these advancements, the application of classification algorithms with DL methods has been limited by the constraints of 1D data, often necessitating disassembly and destructive testing for the root cause analysis. Also, the noise robustness of high-frequency parameter-based methods remains inadequately explored, in spite of their vulnerability to noise. To address these gaps, this study proposes a fault detection and diagnosis method that leverages signal patterns in the full range of operating frequencies to reduce dependency on a single frequency, enhancing robustness to noise and providing richer information for learning algorithms. Our focus is on the reflection coefficient, known for its sensitivity to faults in electronic systems.


\section{Proposed Method}
\label{sec:method}

    \begin{figure}[!t]
		\begin{center}
			\includegraphics[width=\columnwidth]{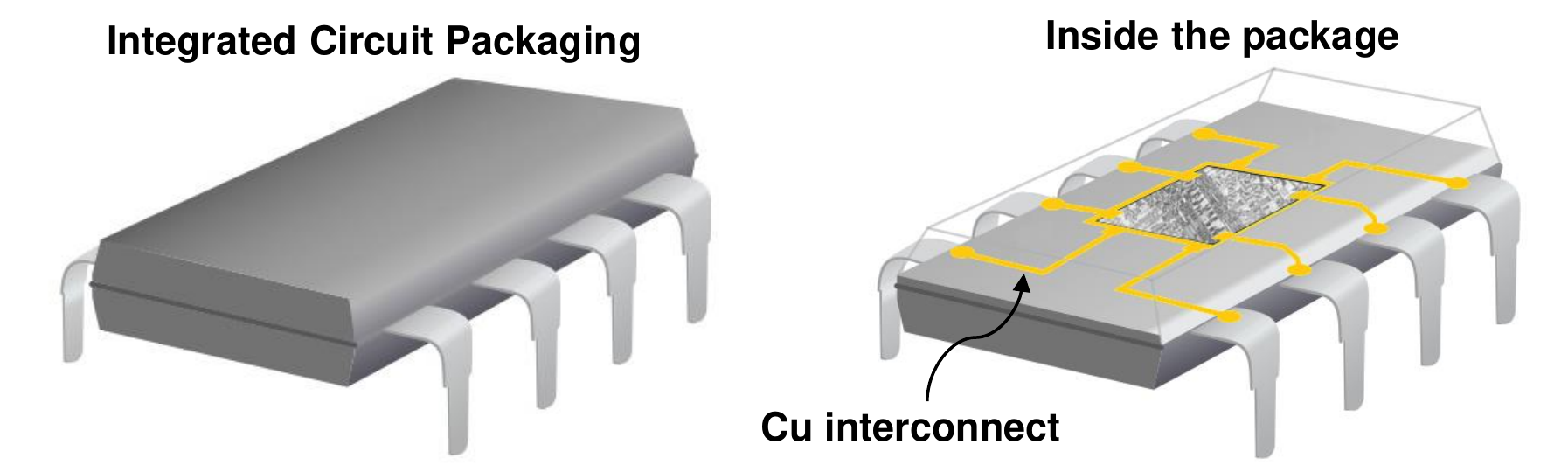}
			\caption{IC packaging and electronic interconnects (Cu interconnects) placed inside the package} 
			\label{fig2}
		\end{center}
    \end{figure}
    \begin{figure}[!t]
		\begin{center}
		\includegraphics[width=\columnwidth]{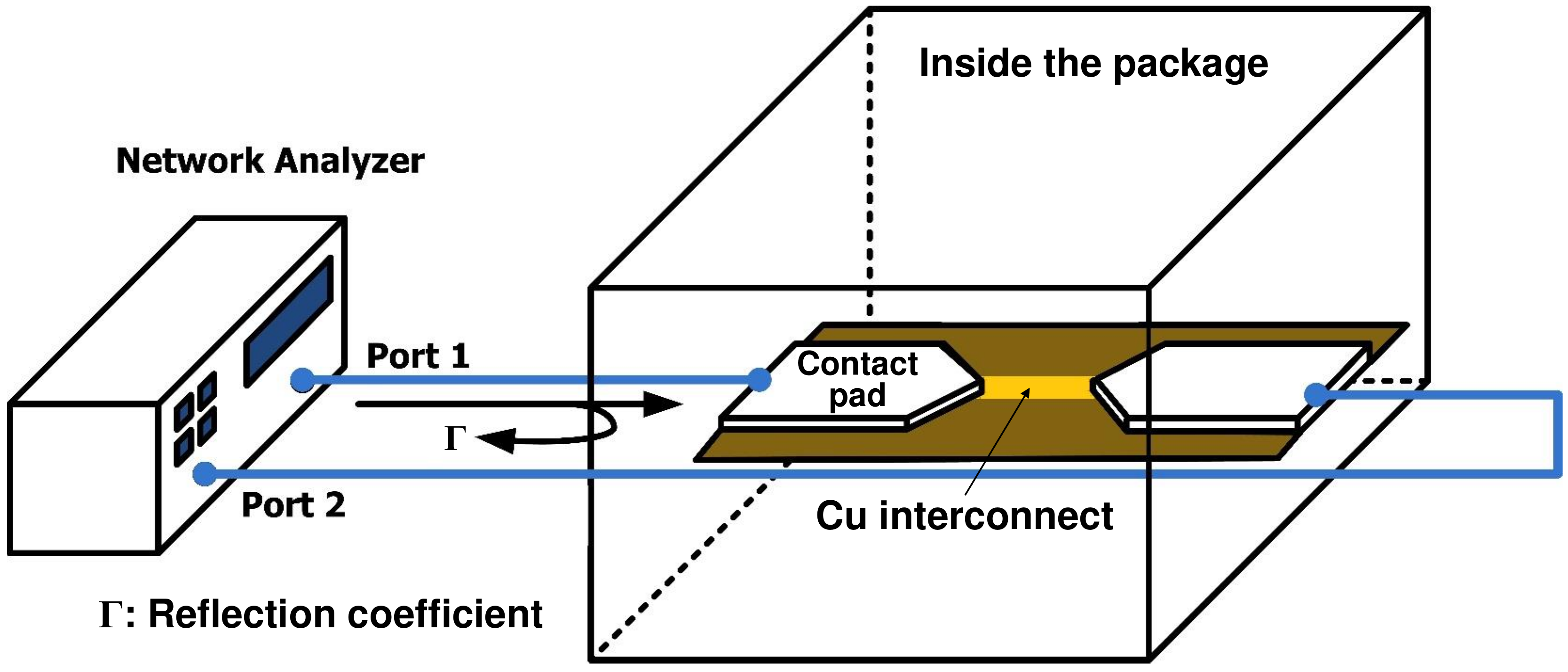}
			\caption{Schematic of the reflection coefficient measurement} 
			\label{fig:measurement}
		\end{center}
	\end{figure} 
    \subsection{Acquisition of Reflection Coefficient Signal Patterns}
        As depicted in \cref{fig2}, the interconnects are placed inside the electronic packages. Thus, in order to investigate the defects that occur in the interconnects, disassembly of the packages has been required so far. However, our proposed method utilizes electronic signals, namely the reflection coefficient, obtained outside the packages to implement non-destructive diagnosis. As shown in \cref{fig:measurement}, the reflection coefficient of an interconnect can be measured by a vector network analyzer at contact pads that are fabricated to be interfaces between the interconnect and the network analyzer. The reflection coefficient, $\Gamma$, is expressed as follows \cite{def_coeff}. 

        \begin{equation}  
                \Gamma = \frac{V^{-}}{V^{+}} = \frac{Z_l-Z_0}{Z_l+Z_0}
            \label{eq:ports}
        \end{equation}      
        where $V^{-}$, $V^{-}$, $Z_l$, $Z_0$ are incident, reflected voltage waves and load, characteristic impedance respectively. The reflection coefficient is used to define the reflected wave with respect to the incident wave. It can be obtained in time or frequency domain, and we compared both in this study. Also, as implied by the equation, the reflection coefficient is a direct indication of the impedance discontinuity. Basically, the reflection coefficient is obtained in the frequency domain. The coefficient captures the stimulus-response waveforms that contain the behavioral models of the interconnects, encompassing resistance, capacitance, inductance, and changes in electrical properties resulting from physical damage. However, the signal patterns of the electrical parameters, along with their relationship to defect states, have not been thoroughly investigated while the time domain characteristics of the parameters have been widely utilized in fault detection and diagnosis \cite{5,tdr_new,tdr_new2,tdr_new3,tdr_new4}. Instead, RF engineers have usually discarded irregularities in signal patterns caused by the defects. In this study, we leverage the reflection coefficient in the frequency domain, which provides valuable two-dimensional (2D) pattern information beyond single-point measurement at a designated time and operating frequency. By identifying specific patterns that correspond to different defect causes and severity levels, we can develop ML and DL algorithms capable of simultaneously detecting defects and providing information on root causes. The signal patterns are fully exploited throughout this study for reliability assessment on electronic interconnects. To the best of our knowledge, this is the first investigation into root cause analysis using electronic signals in this context.       

    \subsection{Severity Rating Ensemble Learning (SREL) Approach for Fault Diagnosis of Electronic Interconnects}

         \begin{figure*}[!t]
    		\begin{center}
    			\includegraphics[width=\linewidth]{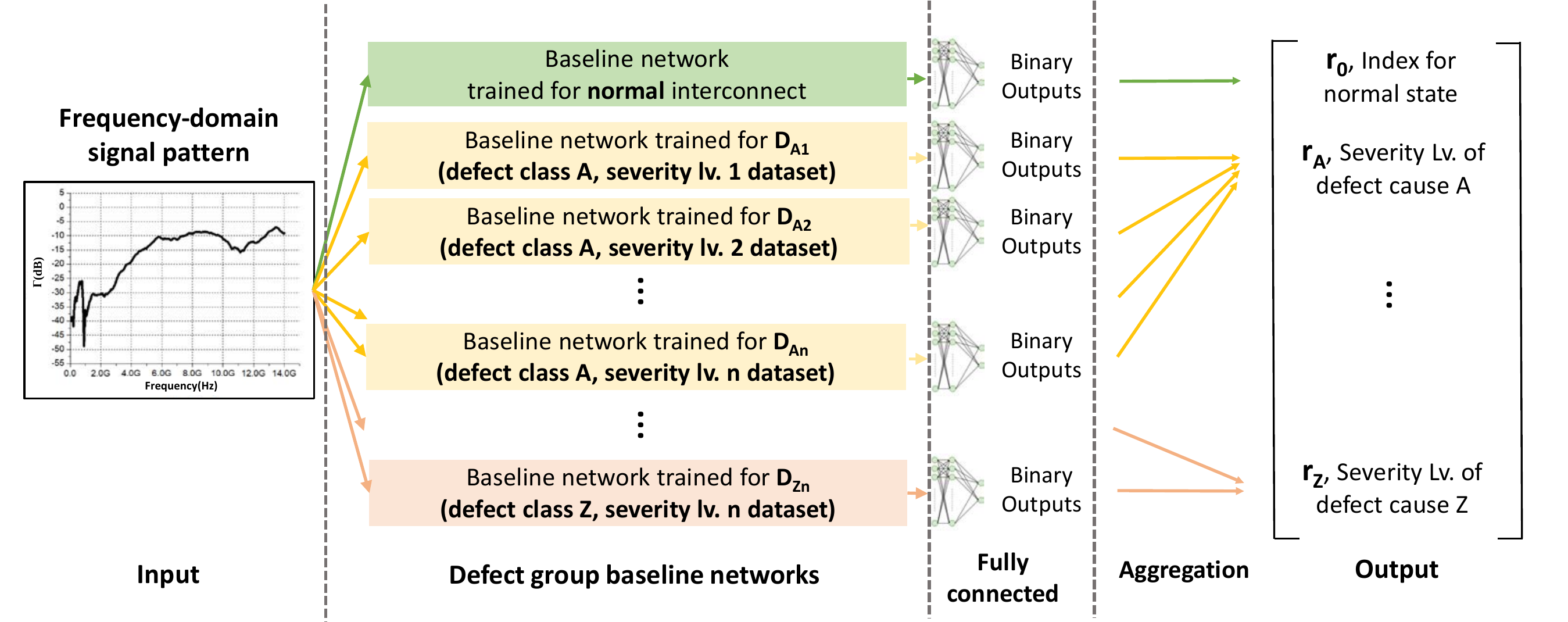}
    			\caption{Structure of the proposed SREL approach for fault diagnosis utilizing signal patterns of interconnects} 
    			\label{fig:overview}
    		\end{center}
	\end{figure*}
        To accurately and robustly extract information regarding the defect from the signal patterns of the reflection coefficient, we propose a severity rating ensemble learning (SREL) approach to determine both the root causes and severity levels. Note that it is also the first research work to apply even the conventional AI algorithms to the electronic signal patterns in terms of fault diagnosis for the interconnects. The SREL is specialized in learning the signal patterns that vary distinctly according to root causes and then subtly to severity levels of defects. As shown in \cref{fig:overview}, an input pattern is fed to the proposed SREL model that is trained with signal pattern datasets gathered according to various defect states. The proposed algorithm contains a series of baseline networks that can utilize a variety of learning models. Subsequently, the baseline networks are tuned with the ordinal severity and cause labels through supervised learning. The training and test data are obtained through experiments and labeled with the causes of defects and their severity levels.

\begin{figure*}[!t]
    		\begin{center}
    			\includegraphics[width=\linewidth]{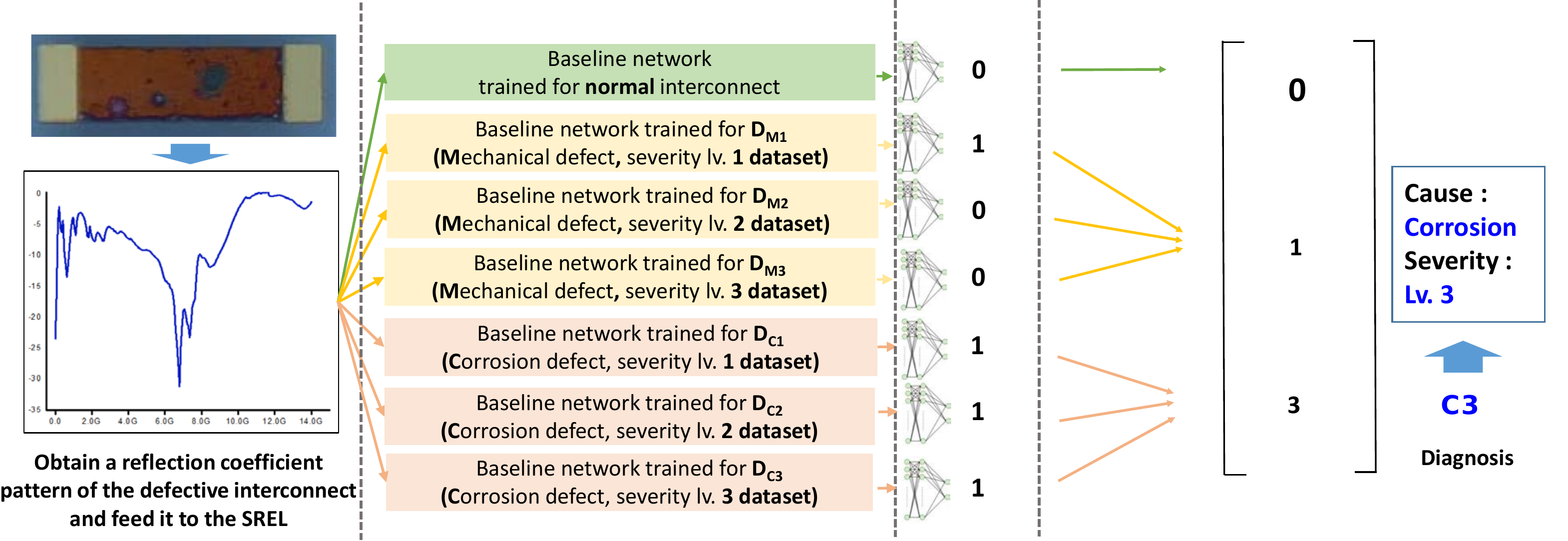}
    			\caption{Working principle of the SREL approach used in our study to distinguish mechanical and corrosion defects with various severity levels} 
    			\label{fig:example}
    		\end{center}
	\end{figure*}

        For example, the label A1 indicates the reflection coefficient signal pattern of an interconnect with the defect cause A and a severity level of 1. Users can adjust the number of causes and severity levels according as their applications.  To train each baseline network, the entire dataset $D$ is divided into two subsets, with severity levels higher or lower than the target severity level. The final fully connected layer computes the probability that the input belongs to the target class using the logistic function. For example, the baseline network to recognize the normal state is trained with the two data subsets: the signal patterns of the normal interconnects and those of the defective ones. The binary output of this baseline network is 1 if the input signal pattern is of the normal state, and 0 if defective. We denote the output of the normal baseline network as $r_0$. Regarding a baseline network targeting a certain defect state, the training process is as follows. The training dataset $D$ is divided into two subsets in reference to the cause of the defect $i$ and the severity level $j$:        
        \begin{equation}
            \begin{aligned} 
            D \; \  = \;& D_{ij}^{+} \cup D_{ij}^{-} \:, \\
                D_{ij}^{+}=\;&\{(x_{ij}, 1) | i = \text{Target cause (i.e. cause A),}\\
                & j \ge	k \text{(Target severity level}\} \:, \\
                D_{ij}^{-}=\;&\{(x_{ij}, 0) | (i=A, j<k) \cup (i=A^c)\}                
            \end{aligned}
            \label{eq:subsets}
        \end{equation}
        
        where the cause of defect $i \in \{A,B,C…,Z\}$, severity level $j\in\{1,2,3…,n\}$, and $x_{ij}$ is the reflection coefficient vector with the cause $i$ and severity level $j$. Each dataset $(D_{ij}^{+},D_{ij}^{-} )$ is used to train the corresponding baseline network in the proposed SREL. Here, the dataset $D_{ij}^{-}$  includes the signal pattern data for the other causes of defect except for the target class. After training, all baseline networks can output a binary decision, 0 or 1. They denote whether the input reflection coefficient vector belongs to the target class. Given the unknown reflection coefficient pattern $x$, we use the baseline networks to make a set of binary decisions and then aggregate them to predict the severity level regarding the cause of the defect $i$, $r_i$.
        \begin{equation}
            \begin{split}
                r_i = \sum_{j=1}^n [f_{ij}(x) > 0]
            \end{split}
            \label{eq:severity}
        \end{equation}
        where $f_{ij} (x)$ is the output of the baseline network and $[\cdot]$ denotes the truth-test operator, which is 1 provided that the inner condition is true, and 0 otherwise. Thus, the output of the proposed SREL model is in a form $(r_0,r_A,r_B,r_C, \cdots r_Z)^T$. The cause and severity of defects from the output vector can be simultaneously determined as follows:
        \newcommand{\argmax}{\mathop{\mathrm{argmax}}}          
        \begin{equation}
            \begin{split}
                \text{Severity} &= \max{(r_0,r_A,r_B,r_C, \cdots r_Z)}\\
                \text{Cause}    &= \argmax {(r_0,r_A,r_B,r_C, \cdots r_Z)}
            \end{split}
            \label{eq:cs}
        \end{equation}
        where $\max(\cdot)$ obtains the maximum value in the bracket, and $\argmax(\cdot)$ finds the position of the max value in the bracket. When the severity levels of different causes are equal, the priority lies on the cause whose baseline networks produce a higher sum of values from the logistic functions. The severity labels are naturally ordinal, and the signal patterns of interconnects with the same cause of defect share similar features. Compared with the softmax-based multiclass classification trained with the complete dataset, the SREL approach can maintain the relative ordinal relationship within the same defect cause group. In this study, we apply the SREL approach to datasets of normal state, mechanical, and corrosion defects of electronic interconnects. \cref{fig:example} depicts the detailed process of fault detection and diagnosis of a defective interconnect by using the SREL approach in this paper. The signal pattern obtained from the interconnect is fed to the pre-trained SREL network. When the aggregated output is $[0, 1, 3]^T$, the maximum value of the output is 3 at the third argument which corresponds to the corrosion defect. Then, the final diagnosis is that the interconnect is defective with corrosion of severity Lv. 3 (C3).

\section{Experimental Design}
\label{sec:ExperimentalDesign}
    \subsection{Test Vehicles and Measurements}
    \subsubsection{Fabrication of test vehicles}
    
        \begin{figure}[!t]
		\begin{center}
			\includegraphics[width=\columnwidth]{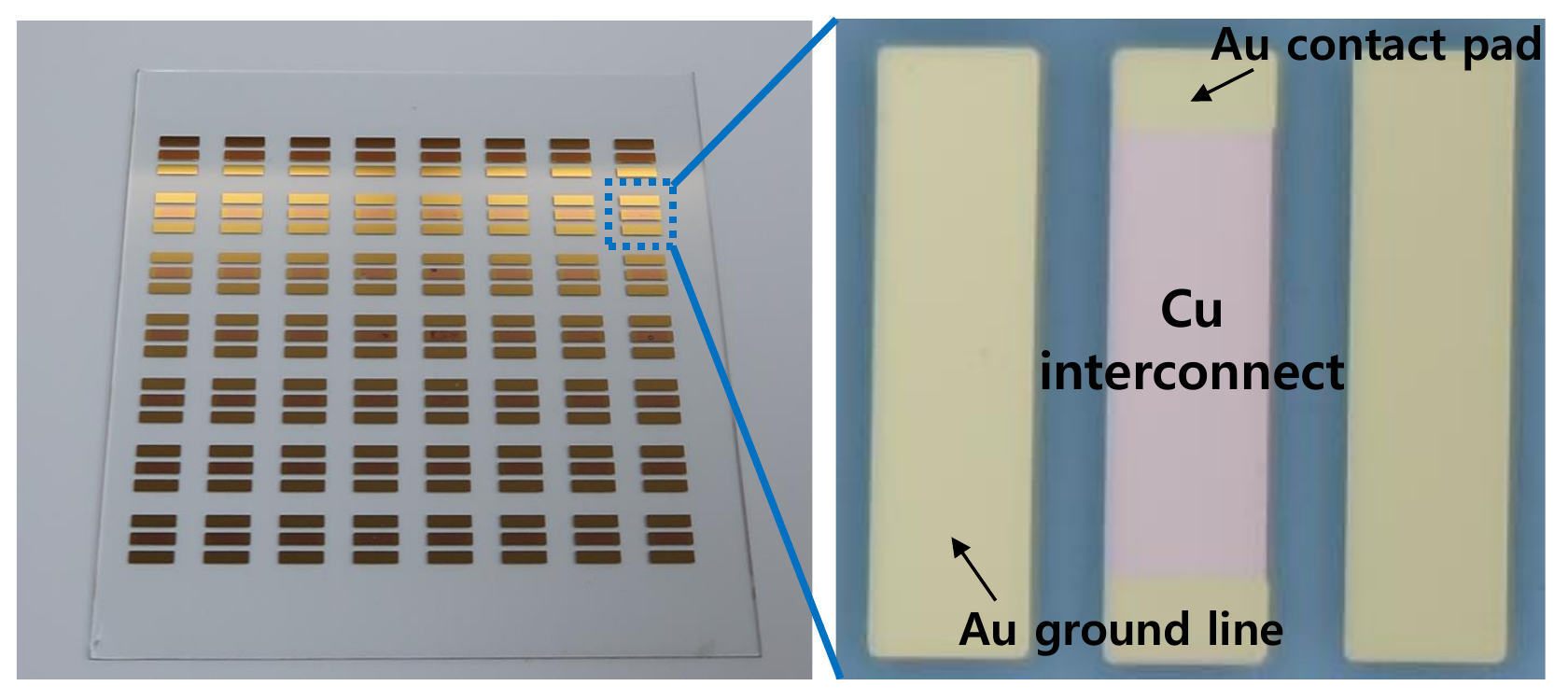}
			\caption{Test batch of electronic interconnect specimens } 
			\label{fig6}
		\end{center} 
        \end{figure}
        
    As a representative form of electronic interconnects, Cu interconnection was used for the target of our proposed fault diagnosis method. For the convenience of experiments and data acquisition, batches of Cu interconnect specimens were fabricated as shown in \cref{fig6}. Gold contact pads were deposited on Cu signal lines. In addition, gold electrodes were fabricated on both sides of the signal line to utilize ground-signal-ground probe tips for the reflection coefficient measurements. The gold pads and electrodes enabled reference points that were unaffected by the environmental stresses. 56 specimens were produced on a glass substrate to obtain as many data as possible with a single batch. In total, 15 batches were used in this study.

        \begin{figure}[t]
		\begin{center}
			\includegraphics[width=\columnwidth]{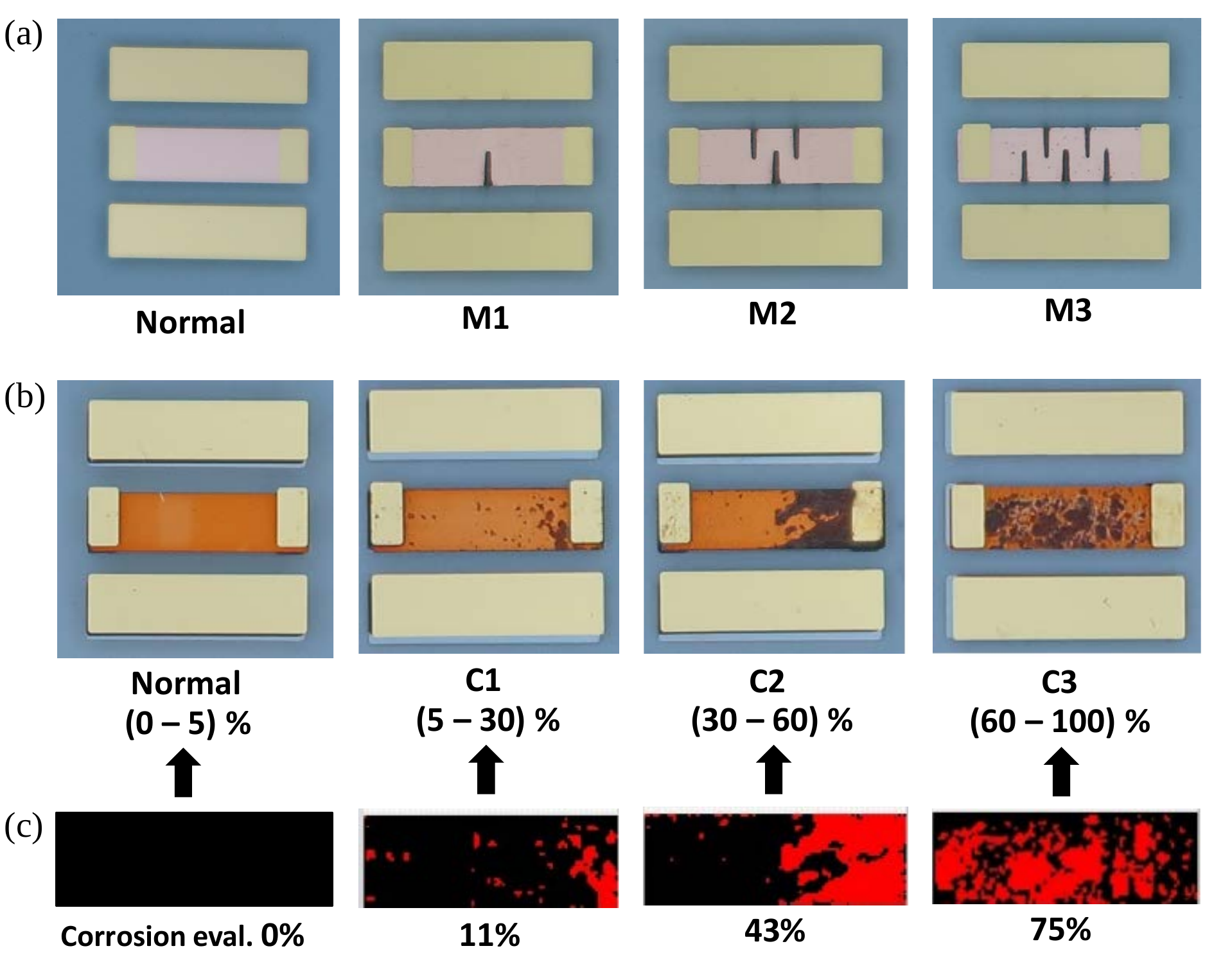}
			\caption{Labeling (a) mechanical and (b) corrosion defects in electronic interconnects. (c) Quantitative evaluation of corrosion using the image processing technique (Note: These images were not utilized for ML and DL techniques)} 
			\label{fig7}
		\end{center}
	\end{figure}
 
    \subsubsection{Inducing and evaluating defects}
    Most electronic part failures are packaging-related, and the electronic packages are susceptible to environmental factors. For instance, thermal expansion produces mechanical stresses that may cause material fatigue, thus leading to crack evolution. Humidity and aggressive chemicals can cause corrosion of the packaging materials. Among the aforementioned failure modes described above, we assumed the two representative root causes: Mechanical and corrosion defects. In addition, regarding the severity level, we classified the defects into four levels: the normal state, Lv. 1 (defective but still usable), Lv. 2 (highly recommended for replacement), and Lv. 3 (out of order). As examples of mechanical defects in the interconnect, we precisely induced 1 mm long and 10 $\mu$m wide cracks in our specimens with a laser cutting machine. The specimens with 1, 3, and 5 cracks were labeled as M1, M2, and M3, respectively, which represent the mechanical defect levels of severity, as depicted in \cref{fig7} (a). To produce corrosion defects in the interconnect specimens, the batches were exposed to the environmental profile according to the MIL-STD-810G humidity method. The environmental conditions were provided by a temperature and humidity chamber (ESPEC). The specimens were photographed every 12 hours. The interconnect specimens with corrosion of (5-30) \%, (30-60) \%, and (60-100) \% were classified as C1, C2, and C3, respectively. \cref{fig7} (b) shows the result of evaluating and labeling corrosion defects in the electronic interconnects. The corrosion defects were quantitatively evaluated using an image processing technique as described in \cref{fig7} (c). Note that images in \cref{fig7} were not utilized for ML and DL techniques, but the reflection coefficient patterns measured from the specimens were fed to the learning algorithms.      

        \begin{figure}[t]
		\begin{center}
			\includegraphics[width=\columnwidth]{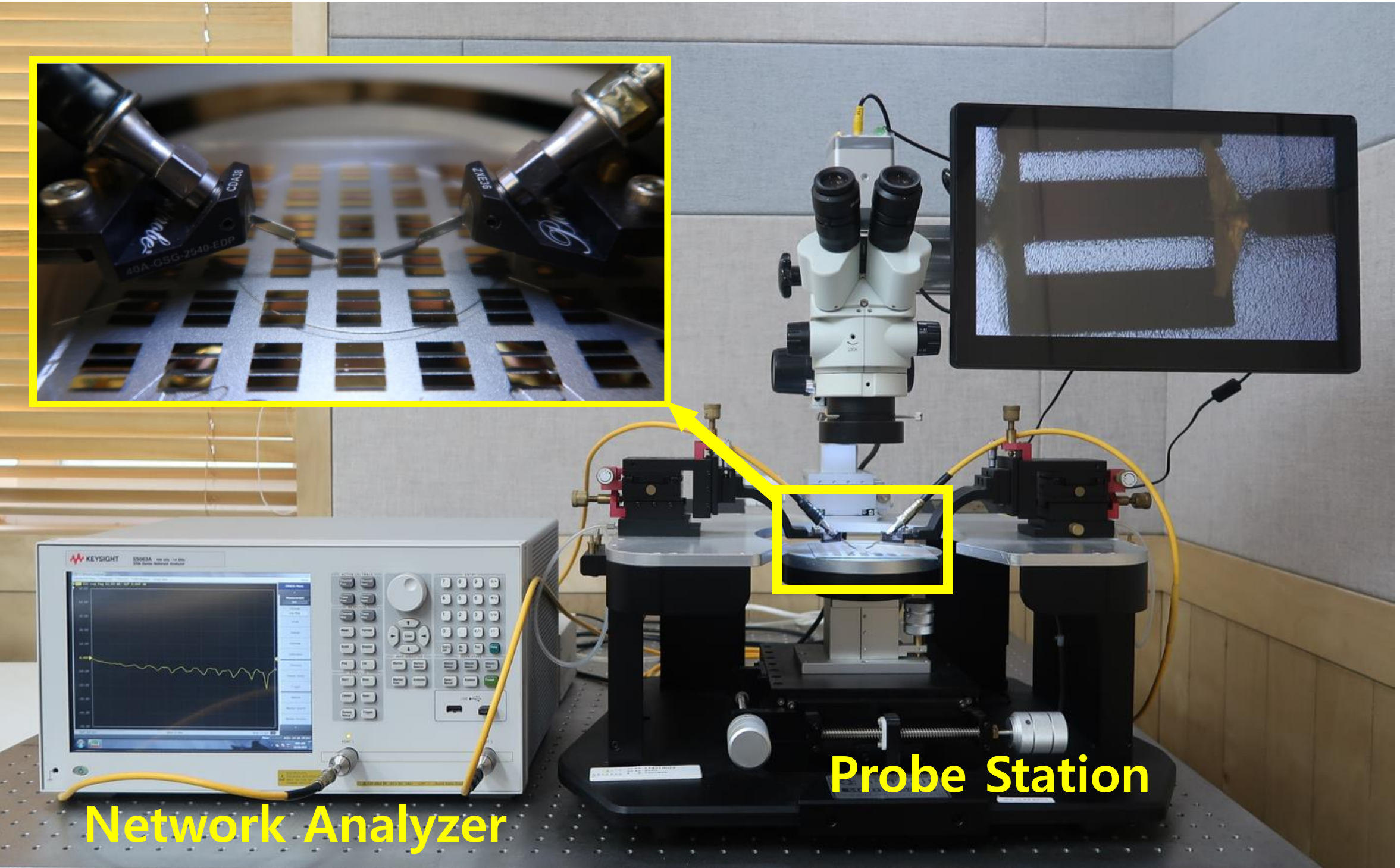}
			\caption{Measurement setup for obtaining the reflection coefficient patterns} 
			\label{fig8}
		\end{center}
    \end{figure} 
   
    \subsubsection{Electronic signal measurements: DC resistance, TDR, S-parameters, and reflection coefficient patterns}
    \cref{fig8} shows an experimental setup for the RF electronic signal measurements used to investigate the reflection-coefficient-pattern indication of various defects in electronic interconnects. With this setup, TDR and S-parameters can be obtained as well. The batch-type test vehicle contained 56 specimens in total. It allowed convenient measurement by using a probe station and helped the specimens be exposed uniformly to the environmental stress. As can be seen in \cref{fig8}, the test vehicle was placed on the probe station (MSTECH Model 5500), and the RF signals were measured. This ex-situ measurement setup helped avoid the degradation of RF cables and connectors, thus concentrating only on the electronic signals related to defect evolution in the interconnect. Both ends of the specimen were contacted by high-frequency probes (ground-signal-ground type, GCB Industries Inc. 40A-GSG-2540-EDP) connected to each port of a vector network analyzer (KEYSIGHT E5063A), thus comprising a two-port network. To investigate the signal-pattern indication of various causes and severity levels of defects in the interconnects, we focused on the change in the reflection coefficient that represents the signal returned to the incident port from the interconnect. The network analyzer also provided a time domain function, which mathematically transforms waveforms in the frequency domain into waveforms in the time domain. Specifically, we used the time domain transformation function of the E5063A analyzer to obtain TDR values. To acquire S-parameters at a designated frequency, we measured the S11 parameters at 8 GHz. In addition, the DC resistance between the ends of the specimen was measured by using a digital multimeter (Fluke 1587 FC). Overall, the reflection coefficient patterns were compared with the electronic signals for the previous fault detection methods such as DC resistance, TDR, S-parameters at a designated frequency. \cref{tab:methods} summarizes the fault diagnosis methods using the electronic signals evaluated in this study.

\begin{table}[]
\centering
    \caption{Comparison of fault Diagnosis methods using electronic signals in this study}
    \label{tab:methods}
\resizebox{\columnwidth}{!}{%
\begin{tabular}{cccc}
\hline
Methods       & Operating frequency(GHz) & Domain & Dimension of information                                    \\ \hline
DC resistance & -                        & -      & \begin{tabular}[c]{@{}c@{}}1D\\ (Single value)\end{tabular} \\
TDR           & 0 - 14                   & Time   & \begin{tabular}[c]{@{}c@{}}1D\\ (Single value)\end{tabular} \\
S-parameter &
  \begin{tabular}[c]{@{}c@{}}8\\ (Designated frequency)\end{tabular} &
  Frequency &
  \begin{tabular}[c]{@{}c@{}}1D\\ (Single value)\end{tabular} \\
\begin{tabular}[c]{@{}c@{}}Reflection coefficient pattern\\ (Our method)\end{tabular} &
  0 - 14 &
  Frequency &
  \begin{tabular}[c]{@{}c@{}}2D\\ (Magnitude-frequency pattern)\end{tabular} \\ \hline
\end{tabular}%
}
\end{table}
    
    \subsection{Comparison Models}
    To evaluate the proposed method with the dataset we collected, we considered nine methods for the classification of causes and severity levels of the defects. These methods can be categorized into three general methods: our SREL approach, multiclass-CNN, and ML. Three types of CNN baseline networks, EfficientNet \cite{21}, 1D-CNN \cite{22,23,24} with 3 layers (1DCNN-3) and 1 layer (1DCNN-1) respectively, were tested with SREL and multiclass-CNN. The multiclass-CNN method is a conventional DL technique for classification using the softmax function. In addition, two conventional ML methods, random forest, and k-mean clustering were evaluated to contrast the performance of DL and ML methods with additive noise.  

    \begin{table}[!t]
    \centering
    \caption{The number of training and test datasets}
    \label{tab:datasets}
    \resizebox{\columnwidth}{!}{
    \begin{tabular}{cccc}
    \hline
    Defect Class & Training Data & Validation Data & Test Data \\
    \hline    
    Normal & 105 & 35 & 35 \\
    M1 & 54 & 18 & 18 \\
    M2 & 48 & 16 & 16 \\
    M3 & 60 & 20 & 20 \\
    C1 & 54 & 18 & 18 \\
    C2 & 57 & 19 & 19 \\
    C3 & 96 & 32 & 32 \\
    \hline
    \end{tabular}
    }
    \end{table}  
    
    \subsection{Data Description and Implementation Details}

    In our experiments, the signal-pattern data of the reflection coefficient were extracted from 790 samples of electronic interconnects. The numbers of samples for each class were 175 for Normal, 90 for M1, 80 for M2, 100 for M3, 90 for C1, 95 for C2, and 160 for C3, respectively. The ratio of training, validation, and test data, applied to all classes, was 6:2:2. \cref{tab:datasets} lists the number of the training, validation, and testing data. The same set of samples with multiclass defect labels was used to train the SREL, multiclass-CNN, and ML methods. Based on this combination of data, we evaluated the diagnostic performance of our proposed method and the other compared methods. 
    
    Hyperparameters for the DL algorithms were optimized as follows. We used the binary cross entropy as the loss function to train the neural networks and minimized it using the Adam optimizer with a learning rate of 0.00005 and a batch size of 256. The number of epochs was determined by the early stopping method. For the K-means clustering model, the number of clusters was 7, and the maximum number of iterations was 300. For the random forest algorithm, the splitting criterion was Gini impurity; the maximum depth of the tree was 5; and the minimum number of samples to split was 2. In this study, the diagnostic performance was evaluated with accuracy defined as TP/(TP+FN), denoting the number of true positives, false positives, true negatives, and false negatives as TP, FP, TN, and FN, respectively. In addition, we examined the macro F1 score to comprehensively evaluate model performance.  

    All methods were implemented with an Intel® Core™ i5-9600k CPU (3.70 GHz), 32 GB RAM, and NVIDIA GeForce RTX 2070 Super GPU. 

\section{Experimental Results and Discussion}
\label{sec:experimentalResults}
    \subsection{Measurement Results: DC resistances, TDR, S-parameters at a Designated Frequency and Signal Patterns of Reflection Coefficient}

   \begin{figure}[!t]
		\begin{center}
			\includegraphics[width=\columnwidth]{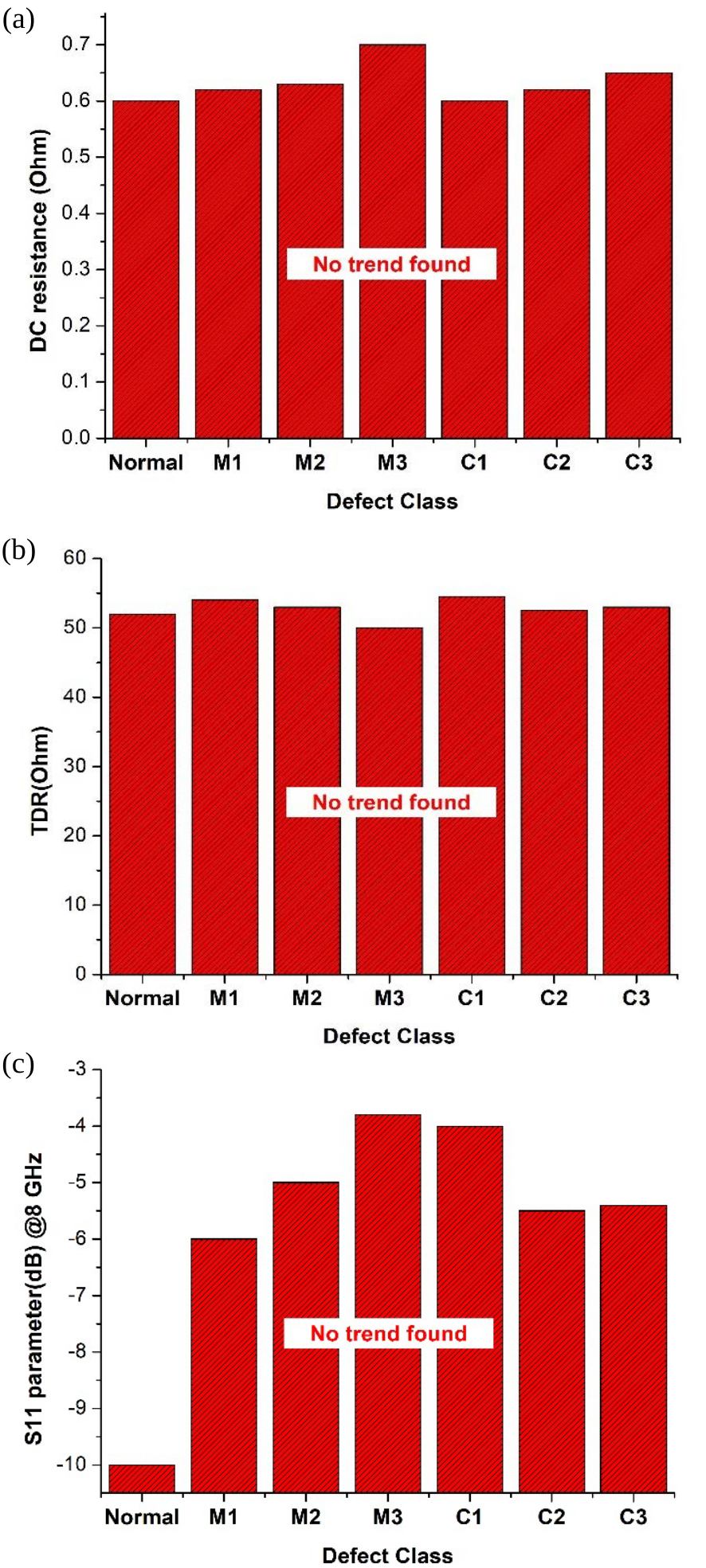}
			\caption{(a) DC resistance, (b) TDR, (c) S-parameter values at a designated frequency according to the defect states} 
			\label{fig9}
		\end{center}
        \end{figure}        

    \begin{figure}[!t]
		\begin{center}
			\includegraphics[width=\linewidth]{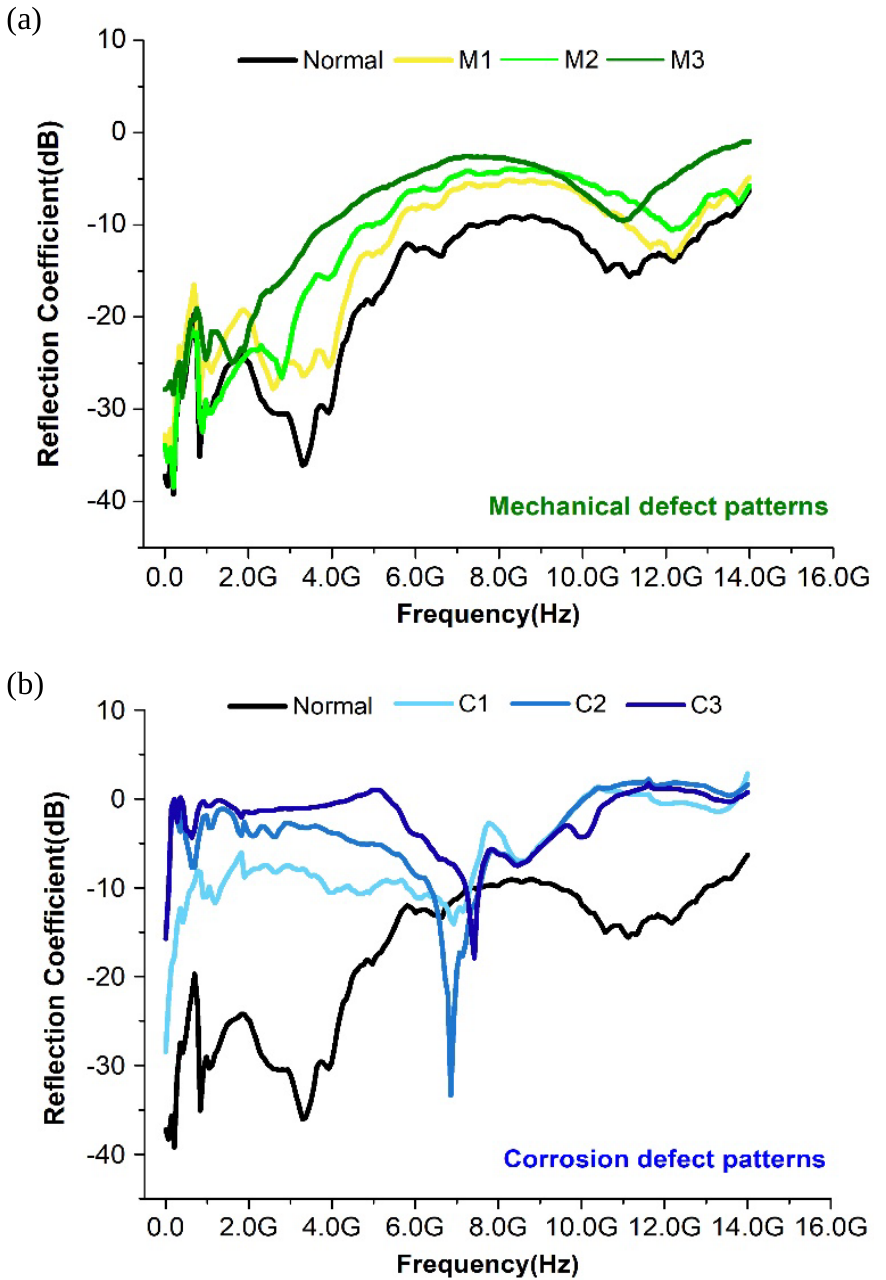}
			\caption{Reflection coefficient patterns of electronic interconnects according to severity levels of (a) mechanical and (b) corrosion defects} 
			\label{fig10}
		\end{center}
    \end{figure}
   
\cref{fig9}(a) shows the DC resistances relative to different defect classes. As the number of cracks in the interconnects increased from 0 to 5, i.e., from the Normal to M3 class, the change in DC resistance was negligible. The average DC resistances for Normal, M1, M2, and M3 were 0.60, 0.62, 0.63, and 0.70~$\Omega$, respectively. Although the values seemed to increase, the changes remained within the range of standard deviation. Similarly, as the corrosion defects progressed, the changes in DC resistances were not significant. The average DC resistances for Normal, C1, C2, and C3 were 0.60, 0.60, 0.62, and 0.65~$\Omega$, respectively. This negligible change in DC resistances across defect states indicates the method’s inability to detect early defects and perform root cause analysis.

As shown in \cref{fig9}(b), the average TDR values corresponding to defect states were 53.2, 58.9, 57.2, 48.8, 59.1, 53.8, and 53.1~$\Omega$ for Normal, M1, M2, M3, C1, C2, and C3, respectively. The average S11 parameter values at 8 GHz, as depicted in \cref{fig9}(c), were -10.0, -6.0, -4.9, -3.7, -4.0, -5.5, and -5.4~dB for Normal, M1, M2, M3, C1, C2, and C3, respectively.  There is no significant variation in TDR values, and no clear trend was observed in the S-parameters at the designated frequency with respect to different defect states. Moreover, it is unfeasible to determine the causes of defects using 1D information, such as DC resistances, S-parameters at a specific frequency, or TDR values, which have been employed by previous fault diagnosis methods for interconnects.

    Meanwhile, the signal patterns of the reflection coefficient provided the ability to distinguish the severity and cause of faults in electronic interconnects. As depicted in \cref{fig10} (a), the reflection coefficient showed identifiable patterns for the mechanical defects. Similarly, \cref{fig10} (b) shows the signal patterns according to the occurrence and evolution of corrosion defects. The graphs are averaged reflection coefficients in the frequency domain among the samples with the same defect state. As the degree of corrosion aggravated from Normal to C1, C2, and finally, to C3, the reflection coefficients of corroded Cu interconnects also showed distinguishable patterns for the corrosion. Overall, defective interconnects exhibited distinguishable reflection coefficient patterns according to the characteristics of defects, proving the capabilities of both early detection and root cause analysis. Interconnects with the same cause of the defects exhibited similar patterns (features of root causes), and the severity levels induced minor changes such as magnitude offsets and peak shifts (features of severity levels). 

    \begin{figure}[!t]
		\begin{center}
			\includegraphics[width=\columnwidth]{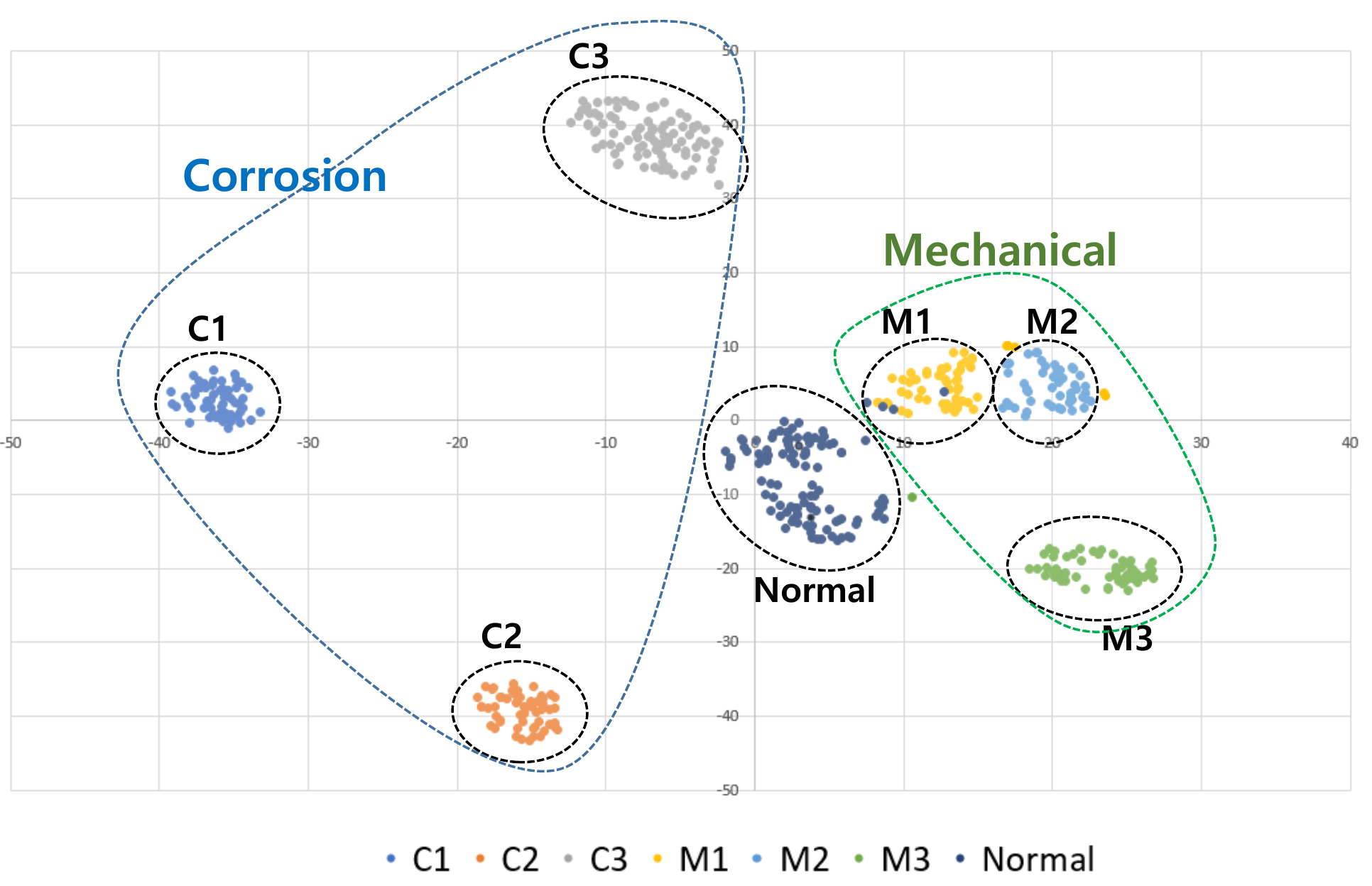}
			\caption{t-SNE dimension reduction results with the reflection coefficient data} 
			\label{fig11}
		\end{center}
    \end{figure}

    \subsection{Visualization of Reflection Coefficient Patterns by t-SNE method}
   
    Measured signal patterns were visualized after dimension reduction in order to examine and understand the feasibility of the signal patterns of reflection coefficient as inputs to ML and DL methods qualitatively. The dimension reduction was conducted based on the t-distributed stochastic neighbor embedding (t-SNE) method that maps high-dimension data into the low-dimensional embedded space \cite{25}. Using t-SNE, we were able to lower the dimension of the reflection coefficient patterns and depict them in a two-dimensional plane. t-SNE is one of the dimension reduction methods to show, herein, how effective reflection coefficient patterns are in distinguishing the root causes and severity levels of defects. As a result, \cref{fig11} contains the visualization of the signal pattern data segmented according to the labeled classes. Based on those t-SNE-based plots, the efficacy of the reflection coefficient patterns for diagnosing the cause/severity of defects could be verified. In other words, by observing the well-clustered pattern data in the reduced dimension, we were able to confirm that the reflection coefficient patterns are effective data for learning algorithms. Therefore, based on the proposed method, high diagnostic accuracy could be obtained by learning the signal patterns.

    \subsection{Diagnostic Performance}
    \subsubsection{Diagnostic performance on data without noise}

    \begin{table}[!t]
    \centering
    \caption{Diagnostic performance of comparison methods
    }
    \label{tab:performance}
    \resizebox{\columnwidth}{!}{
    \begin{tabular}{ccccc}
    \hline
    \multirow{2}{*}{Model} & \multirow{2}{*}{Accuracy} & Macro & \# of & \multirow{2}{*}{Inference} \\
    & & F1-Score & parameters & \\
    \hline
    SREL  & \multirow{2}{*}{99.3\,\%} & \multirow{2}{*}{0.991} & \multirow{2}{*}{63.7} & \multirow{2}{*}{30.1\,ms} \\
    (EfficientNet) & & & & \\
    SREL & \multirow{2}{*}{97.2\,\%} & \multirow{2}{*}{0.970} & \multirow{2}{*}{4.7} & \multirow{2}{*}{3.88\,ms} \\
    (1DCNN-3) & & &  & \\
    SREL  & \multirow{2}{*}{96.5\,\%} & \multirow{2}{*}{0.961} & \multirow{2}{*}{1.2} & \multirow{2}{*}{2.62\,ms} \\
    (1DCNN-1)& & &  & \\ \hline
    Multiclass-CNN & \multirow{2}{*}{98.6\,\%} & \multirow{2}{*}{0.979} & \multirow{2}{*}{63.7} & \multirow{2}{*}{32.7\,ms} \\
    (EfficientNet) & & &  & \\
    Multiclass-CNN  & \multirow{2}{*}{95.1\,\%} & \multirow{2}{*}{0.941} & \multirow{2}{*}{4.7} & \multirow{2}{*}{4.26\,ms} \\
    (1DCNN-3)& & &  & \\
    Multiclass-CNN  & \multirow{2}{*}{88.9\,\%} & \multirow{2}{*}{0.868} & \multirow{2}{*}{1.2} & \multirow{2}{*}{3.53\,ms} \\
    (1DCNN-1)& & &  & \\ \hline
    ML (RF) & 98.6\,\% & 0.979 & - & 7.36\,ms \\
    ML (K-means) & 81.2\,\% & 0.821 & - & 2.56\,ms \\
    \hline
    \end{tabular}
    }
    \end{table}
    
    The diagnostic performance of the proposed and the other methods on electronic interconnect data are listed in \cref{tab:performance}. First, the SREL approach with the EfficientNet baseline network achieved a diagnostic accuracy of 99.3\%, outperforming the other models. In the case of the 1D-CNN baseline network, the 3-layer network showed better performance than the 1-layer network. Second, the multiclass CNN model with the EfficientNet backbone predicted the cause/severity of defects with a diagnostic accuracy of 98.6\%. The diagnostic performance of CNN-based models can be further improved by stacking the networks deeper. Finally, random forest, another ensemble technique based on ML, produced a diagnostic performance of 98.6\%, indicating that the signal patterns of the interconnects were also effective features for ML techniques. Among the various methods, k-means clustering had the lowest accuracy of 81.2\%. Furthermore, we presented the number of parameters and inference time of DL methods. Notably, the SREL architecture provided faster inference as well as a smaller number of parameters compared to the multiclass DL method with the same backbone network. It is also the first research work to apply even the conventional learning techniques to signal patterns in terms of fault diagnosis for electronic packages. Hence, meaningful is that the conventional ML and DL methods work well with the reflection coefficient patterns that we experimentally gathered according to the causes and severity levels of defects.

    \subsubsection{Diagnostic performance on data with additive noise}
    
\begin{figure}[!t]
		\begin{center}
			\includegraphics[width=\linewidth]{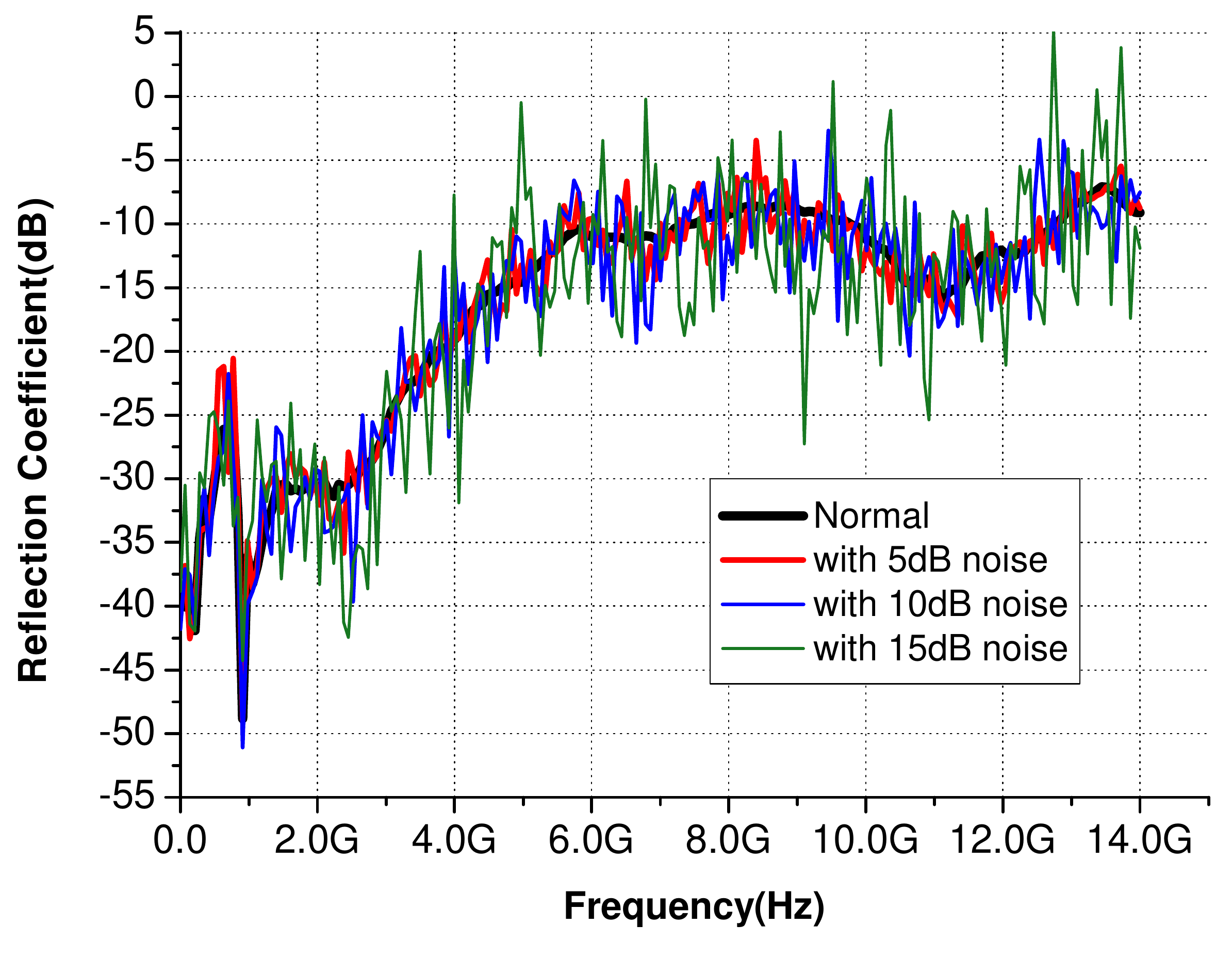}
			\caption{Shapes of the reflection coefficient patterns with increasing additive noise levels} 
			\label{fig12}
		\end{center}
	\end{figure}

\begin{figure}[!t]
		\begin{center}
			\includegraphics[width=\linewidth]{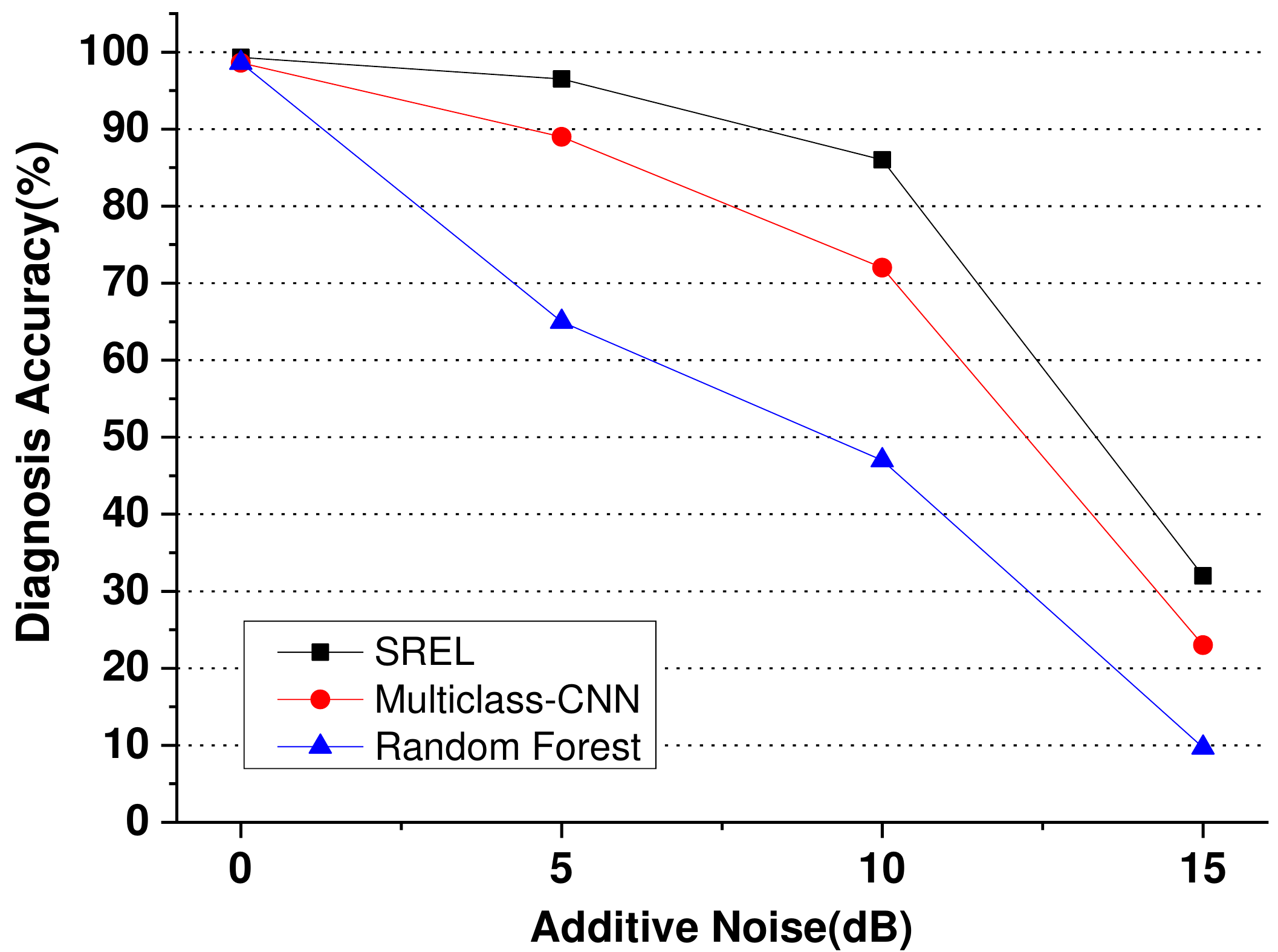}
			\caption{Diagnostic performance of the compared methods at increasing noise levels} 
			\label{fig13}
		\end{center}
	\end{figure}

    Fault diagnosis methods aim to enhance the reliability of electronics in real-world industrial applications \cite{26}. An issue in such sites is noise resulting from environmental and operational randomness including interferences of electrical sensors and devices \cite{27}. This industrial noise can be simulated by using white Gaussian noise \cite{28}. Thus, in this study, we tested the conventional AI algorithms and our proposed model with the data injected with the white noise as described in \cref{fig12}. The noise was produced using MATLAB with the white Gaussian noise function, which provides Gaussian power noise. In results, all methods showed lower diagnostic accuracy as the additional noise increased. This was because the noise caused variation in the characteristics and distributions of the signal data. Notably, the proposed SREL model outperformed the other methods at all levels of white noise tested herein, whereas the performance of the other methods rapidly degraded with the noise. The diagnostic results on the test data with additive noise are shown in \cref{fig13}. In summary, the proposed method exhibited more robust and stable performance against additive noise compared with conventional ML and the multiclass-CNN methods. ML methods, especially the random forest, showed faster inference with fewer parameters even compared to the SREL methods, but the ML methods turned out to be extremely vulnerable to noise.

    \subsection{Discussion}

    By gathering the reflection coefficient patterns of electronic interconnects according to causes and severity levels of defects, we found out that the signal patterns of reflection coefficients possessed the ability of fault diagnosis, especially with root cause analysis. Dimension reduction using t-SNE clearly showed that the patterns were well grouped according to the defect information, hence indicating that the patterns were effective input data for ML and DL methods. Accordingly, conventional ML and basic DL techniques performed fault diagnosis on the electronic interconnects with satisfying accuracies, which was also a novel result in fault detection and diagnosis of electronic packages because the signal patterns had not been utilized in the field. Although the conventional ML and basic DL methods showed good diagnostic results with the patterns of the reflection coefficient without noises, the industrial noises would deteriorate the diagnostic performance. The diagnostic results on data with additive noise indicated that the ML methods were more vulnerable to noise than DL, confirming that DL was more robust to industrial noises than ML \cite{29}. In this study, our SREL model showed better performance with or without noise, hence proving the capability of a more accurate and robust fault diagnosis method for electronic packaging. The excellent performance of SREL results from the approach where SREL divides a defect cause and severity problem into a series of binary classification sub-problems. It obtains estimates by aggregating the results of each sub-problem. At this point, the final estimation error is bound by the maximum error of binary estimators, as mathematically proven by Chen et al. \cite{30}. Also, it is known that the binary output aggregation outperforms the softmax-based multiclass-classification methods \cite{30}. In this study, the softmax-based multiclass CNN methods failed to account for the ordinal relationships between defect severity levels that the signal patterns showed. Thus, instead of multiclass-CNN methods with a softmax classifier, the proposed method is preferred for the defect cause and severity analysis. 

    The guideline for applying our method is as follows. Reflection coefficients should be obtained with regard to defect states to utilize the SREL approach in industrial applications. After setting the ground truth for the learning algorithm, the SREL model can be built and trained. Then, the trained network could be deployed to industrial fields. When users monitor the signal patterns of the reflection coefficient obtained from components of interest (irrespective of whether this step is operated regularly or not), they can feed the pattern to the network. Then, the SREL model can determine the root cause and severity of the defect based on the pre-trained network. With this information regarding the defects, users would respond to the situation early and effectively.

The magnitude of the S-parameter may vary slightly depending on the size of the crack or the shape of the corrosion. However, the S-parameter pattern can effectively distinguish between the causes of defects, whether mechanical damage or corrosion. In reality, these defects evolve continuously rather than discretely. Based on the results of this study, we plan to investigate the feasibility of using S-parameter patterns to estimate the crack size or degree of corrosion.

In addition, we provide a comparative evaluation of the fault diagnosis method using electrical signals, as shown in \cref{fig:comparison}. The comparison was made from five perspectives: cost (ability to detect defects early enough to ensure sufficient remaining useful life), practicality (ease of implementation and a wide range of applications), cost (implementation expenses), noise robustness (stability of results under noise), and root cause analysis (ability to determine the root causes of defects). Compared to other methods, signal-pattern analysis using AI models offers superior noise robustness and root cause analysis capability. Analyzing the signal pattern across the full frequency range reduces dependency on specific frequencies and provides rich information simultaneously. However, the cost of implementing the method can be a challenging issue

        \begin{figure}[!t]
		\begin{center}
			\includegraphics[width=\linewidth]{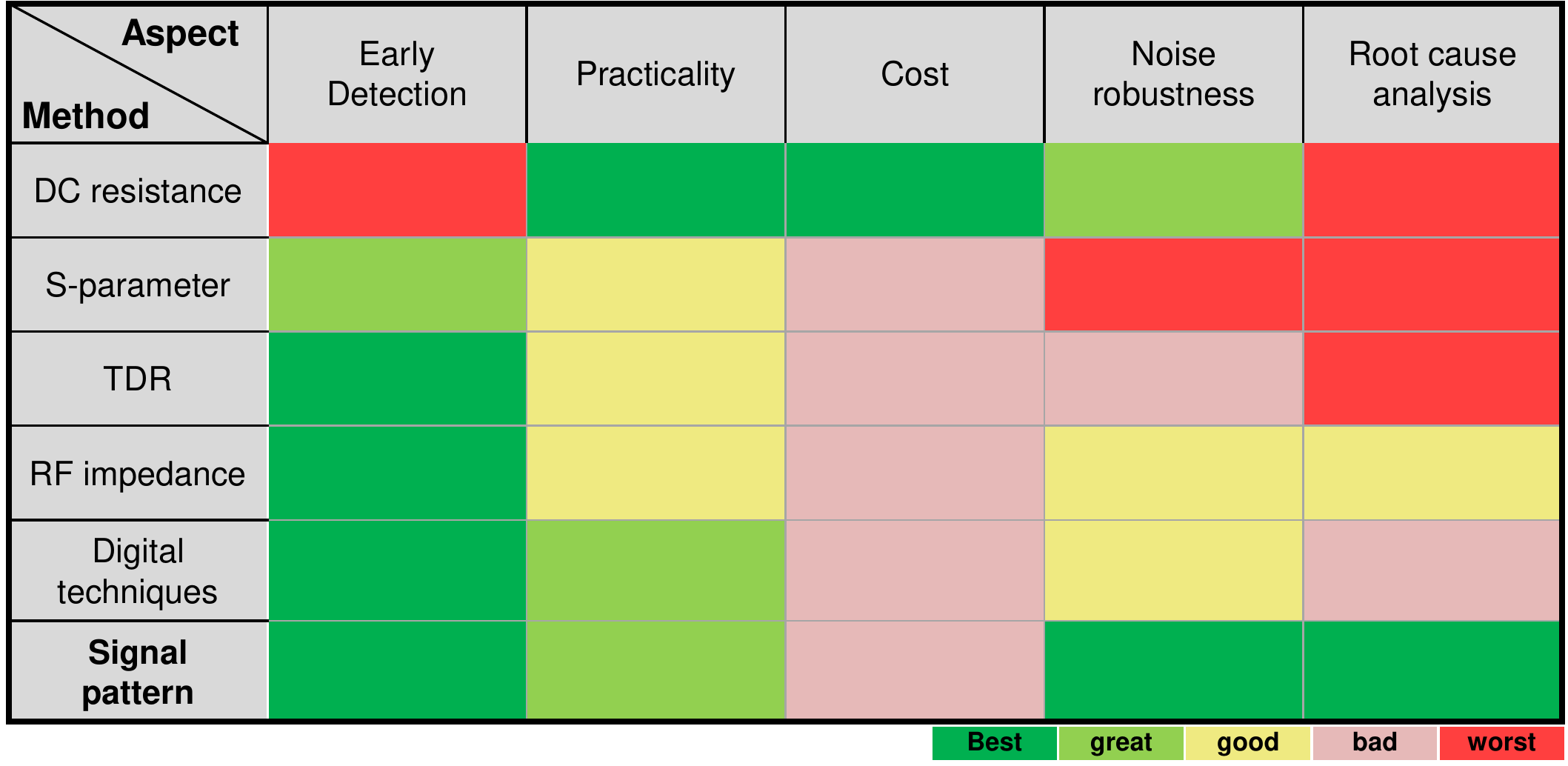}
			\caption{Comparison of the fault diagnosis methods using electrical signals including the signal-pattern analysis} 
			\label{fig:comparison}
		\end{center}
	\end{figure}

\section{Conclusion}
\label{sec:conclusion}
    In this paper, we presented a novel approach for non-destructive fault diagnosis of electronic interconnects, using the reflection coefficient patterns to distinguish the causes and severity levels of defects.
    We focused on corrosion and mechanical defects in electronic interconnects with varying severity levels. In the experimental results, we demonstrated that the reflection coefficient patterns exhibited distinguishable features for 7 states (Normal, M1, M2, M3, C1, C2, C3), enabling root cause analysis and early defect detection in a non-destructive way. On the other hand, it was unable to distinguish the defect causes by using the 1D electronic signals including  DC resistance, TDR, and S-parameter at a designated frequency. 
    The signal patterns of the reflection coefficient provided effective input to both ML and DL techniques, overcoming the limitations of traditional time domain signal analysis. Utilizing existing CNN (1D-CNN, EfficientNet) and ML (RF, K-means clustering) methods, we achieved a maximum diagnostic accuracy of 98.6 \%, indicating the efficacy of the reflection coefficient patterns as features for fault diagnosis.
    To further enhance diagnostic performance and noise robustness, we introduced the SREL method, which fully utilized the unique characteristics of the signal patterns. Our proposed model achieved a maximum diagnostic accuracy of 99.3 \% with our experimental data, outperforming conventional ML and multiclass-CNN approaches, particularly under increased noise level conditions. Our proposed fault diagnosis method facilitates early detection and provides a simultaneous cause and severity analysis, eliminating the need for secondary tools, all while maintaining robustness against noise.
    The potential applications of the proposed method include fault detection and diagnosis in a wide range of electronic interconnects, such as integrated circuits, printed circuit boards, and flexible electronics. In addition, it could be utilized in quality control processes in the manufacturing line to improve product reliability and reduce downtime. 

    In the future, we plan to explore the extension of our work to different types of electronic devices and materials to further validate its applicability and robustness with various real industrial scenarios. Moreover, the integration of real-time monitoring systems and automatic fault diagnosis algorithms based on the proposed method would enable proactive maintenance and minimize potential faults and damages in electronic systems.

\section*{Acknowledgments}
    This study was supported by the research grant of The University of Suwon in 2023.
    
 \bibliographystyle{elsarticle-num} 
 \bibliography{cas-refs}





\end{document}